\documentclass[11pt]{article}
\usepackage{coling2020}
\usepackage{times}
\usepackage{url}
\usepackage{latexsym}
\usepackage{amsmath}
\usepackage{amssymb}
\usepackage{array}
\usepackage{booktabs}
\usepackage{graphicx}

\title{TeRo: A Time-aware Knowledge Graph Embedding via Temporal Rotation}

\author{Chengjin Xu \and Mojtaba Nayyeri \and Fouad Alkhoury \and Hamed Shariat Yazdi\\
  Smart Data Analytics Group, University of Bonn / Germany \\
  {\tt  	\{xuc,nayyeri\}@iai.uni-bonn.de} \\
  {\tt  	s6foalkh@uni-bonn.de} \\
  {\tt  	shariatyazdi@gmail.com} \\\AND
  Jens Lehmann \\
  University of Bonn / Germany \\
  Fraunhofer IAIS/ Germany\\
  {\tt jens.lehmann@iais.fraunhofer.de}\\}

\date{}

\begin{document}
\maketitle
\begin{abstract}
  In the last few years, there has been a surge of interest in learning representations of entities and relations in knowledge graph (KG). However, the recent availability of temporal knowledge graphs (TKGs) that contain time information for each fact created the need for reasoning over time in such TKGs. In this regard, we present a new approach of TKG embedding, TeRo, which defines the temporal evolution of entity embedding as a rotation from the initial time to the current time in the complex vector space. Specially, for facts involving time intervals, each relation is represented as a pair of dual complex embeddings to handle the beginning and the end of the relation, respectively. We show our proposed model overcomes the limitations of the existing KG embedding models and TKG embedding models and has the ability of learning and inferring various relation patterns over time. Experimental results on four different TKGs show that TeRo significantly outperforms existing state-of-the-art models for link prediction. In addition, we analyze the effect of time granularity on link prediction over TKGs, which as far as we know has not been investigated in previous literature.
\end{abstract}

\section{Introduction}~\label{intro}
%
%
\blfootnote{
    
    \hspace{-0.65cm}  
    This work is licensed under a Creative Commons 
    Attribution 4.0 International License.
    License details:
    \url{http://creativecommons.org/licenses/by/4.0/}.
}
In recent years, a number of sizable Knowledge Graphs (KGs) have been constructed, including DBpedia~\cite{Dbpedia}, YAGO~\cite{YAGO}, Nell~\cite{NELL} and Freebase~\cite{Freebase}. In these KGs, a fact is represented as a triple (\textit{s, r ,o}), where \textit{s} (subject) and \textit{o} (object) are entities (nodes), and \textit{r} (relation) is the relation (edge) between them.

Several KG embedding (KGE) models are developed to perform learning and inference over these KGs~\cite{TransE,DISTMULT,ComplEx,RotatE,QuatE}. The most common learning task for these models is link prediction, which is to complete a fact with the missing entity. For instance, one can use a KGE model to perform an object query like (\textit{Barack Obama, visits, ?}). In this case, there are several valid answers to this question, regardless of the time factor. Obviously, the inclusion of time information can make this query more specific, e.g., (\textit{Barack Obama, visits, ?, 2014-07-08}). 

Some temporal KGs (TKGs) including ICEWS~\cite{ICEWS2015}, GDELT~\cite{GDELT}, YAGO3~\cite{YAGO3} and Wikidata~\cite{Wikidata}  store billions of time-aware facts as quadruples (\textit{s, r ,o, t}) where \textit{t} is the time annotation, e.g., (\textit{Barack Obama, visits, Ukraine, 2014-07-08}). 
The availability of these TKGs that exhibits complex temporal dynamics in addition to its multi-relational nature has created the need for approaches that can characterize and reason over them.
Traditional KGE models disregard time information, leading to an ineffectiveness of performing link prediction on TKGs involving temporary relations (e.g., \textit{visits}, \textit{live in}, etc.).

To tackle this problem, TKG embedding (TKGE) models encode time information in their embeddings. Such TKGE models~\cite{KnowEvolve2017,leblay,TA-TransE,TDistMult,HyTE,RE-NET} were shown to have better performances on link prediction over TKGs than traditional KGE models. However, most of the existing TKGE models are the extensions of TransE~\cite{TransE} and DistMult~\cite{DISTMULT}, and thus are not fully expressive for some relation patterns~\cite{RotatE}.

In this paper, we propose a novel approach for TKGEs, TeRo, which defines the temporal evolution of an entity embedding as a rotation from the initial time to the current time in the complex vector space. We show the limitation of the existing TKGE models and the advantage of our proposed model on learning various relation patterns over time.

Specially, for facts involving time intervals, each relation is represented as a pair of dual complex embeddings which are used to handle the beginning and the end of the relation, respectively. In this way, TeRo can adapt well to datasets where time annotations are represented in the various forms: time points, beginning or end time, time intervals. 

Most of previous TKGE-related works as far as we know use specific time granularities for various TKGs. For example, the time granularity of the ICEWS datasets is fixed as 24 hours in~\cite{HyTE,ATiSE}. In this work, we adopt various time-division approaches for different TKG datasets and investigate the effect of the length of the time steps on the performance of our model. 

To verify our approach, we compare the performance of our proposed models on link prediction and time prediction tasks over four different TKGs with the state-of-the-art KGE models and the existing TKGE models. The experimental results demonstrate that our proposed model outperforms other baseline models significantly by inferring various relation patterns and encoding time information.

\section{Related Work}~\label{Related Work}
KGE models can be roughly classified into distance-based models and semantic matching models.

Distance-based models measure the plausibility of a fact as the distance between the two entities, usually after a translation or rotation carried out by the relation. 
A typical example of distance-based models is TransE~\cite{TransE}. TransE exhibits deficiencies when learning 1-n relations. Thus, various extensions of TransE~\cite{TransH,TransD,TransR,TransComplex}, were proposed to tackle this problem. They use different mapping methods to project entities from entity space to relation space. Specially, RotatE~\cite{RotatE} defines each relation as a rotation from the subject to the object. Nevertheless, these distance-based distance models are still unable to capture reflexive relations which can hold, i.e.~for a particular relation $r$ each entity is related to itself via $r$. In distance-based models, the values or the phases of vectors for all reflexive relations are enforced to be 0, which does not allow to fully express the semantic characteristics of these relations. 

Semantic matching models measure plausibility of facts by matching latent semantics of entities and relations embodied in their embedding representations. A few examples of such models include RESCAL~\cite{RESCAL}, DistMult~\cite{DISTMULT}, ComplEx~\cite{ComplEx}, QuatE~\cite{QuatE} and GeomE~\cite{GeomE}. RESCAL and DistMult cannot capture asymmetric relations since the score of the triple $(s,r,o)$ is always equal to the score of its symmetric triple $(o,r,s)$. ComplEx, QuatE and GeomE have been proven to be able to capture various relation patterns for static KGs, but cannot model temporary relations in TKGs due to their ignorance of time information.

Recent research illustrated that the performances of KGE models can be further improved by incorporating time information in TKGs. Some TKGE models are extended from TransE, e.g., TTransE~\cite{leblay}, TA-TransE~\cite{TA-TransE}, HyTE~\cite{HyTE} and ATiSE~\cite{ATiSE}. Another part of TKGE models are temporal extensions of DistMult, e.g., Know-Evolve~\cite{KnowEvolve2017}, TDistMult~\cite{TDistMult} and TA-DistMult~\cite{TA-TransE}. Similar to TransE and DistMult, these TKGE models have issues with capturing reflexive relations or asymmetric relations. Specially, DE-SimplE~\cite{DE-SimplE} incorporates time information into diachronic entity embeddings and has capability of modeling various relation patterns. However, this approach only focuses on event-based TKG datasets, and cannot model facts involving time intervals shaped like [\textit{2003-\#\#-\#\#}, \textit{2005-\#\#-\#\#}].

\section{A Novel TKGE Approach based on Temporal Rotation}~\label{method}
Although various KGE models have been developed to learn multi-relational interactions between entities, all of them have problems with inferring temporary relations which are only valid for a certain time point or a last for a certain time period. To illustrate this by example, assume we are given a quadruple (\textit{Barack Obama, visits, France, 2014-02-12}) as a training sample, where the relation \textit{visits} is a temporary relation. If we query (\textit{Barack Obama, visits, ?, 2014-07-08}), a trained static KGE model probably returns the incorrect answer \textit{France} due to 
the validness of the triple \textit{Barack Obama, visits, France}, while the correct answer is \textit{Ukraine} considering the given time constraint. On the other hand, most of the existing TKGE models, which were extended from TransE~\cite{TransE} and DistMult~\cite{DISTMULT}, incorporate time information in the embedding space, but have limitations on learning transitive relations or asymmetric relations as discussed in Section~\ref{Related Work}.

To overcome the limitations of these existing KGE and TKGE models on learning and inferring over TKGs, we propose a new TKGE model, TeRo, which defines the temporal evolution of an entity embedding as a rotation in the complex vector space. Let $\mathcal{E}$ denote the set of entities, $\mathcal{R}$ denote the set of relations, and $\mathcal{T}$ denote the set of time steps. 
Then a TKG is a collection of factual quadruples (\textit{s}, \textit{r}, \textit{o}, \textit{t}), where $s, o \in \mathcal{E}$ are the subject and object entities, $r\in \mathcal{R}$ is the relation, $t$ denotes the actual time when the fact occurs. For any time $t$, we have a time step $\tau\in \mathcal{T}$ representing this actual time. We map $s$, $r$, $o$ to their complex embeddings, i.e., $\textbf{s},\textbf{r},\textbf{o} \in \mathbb{C}^{k}$; then we define the functional mapping induced by each time step $\tau$ as an element-wise rotation from the time-independent entity embeddings $\textbf{s}$ and $\textbf{o}$ to the time-specific entity embeddings $\textbf{s}_{t}$ and $\textbf{o}_t$. The mapping function is defined as follows:
\begin{equation}
\begin{aligned}
    &\textbf{s}_{t}=\textbf{s}\circ\boldsymbol{\tau}, \quad \textbf{o}_{t}=\textbf{o}\circ\boldsymbol{\tau}
\end{aligned}\label{Equation:entity representation}
\end{equation}
where $\circ$ denotes the Hermitian dot product between complex vectors. Here, we constrain the modulus of each element of $\boldsymbol{\tau}\in\mathbb{C}^{k}$, i.e., $\boldsymbol{\tau}_{j}\in\mathbb{C}$, to be $|\boldsymbol{\tau}_{j}|=1$. By doing this, $\boldsymbol{\tau}_{j}$ is of the form $e^{i\theta_{\tau,j}}$, which corresponds to a counter-clockwise rotation by ${\theta_{\tau,j}}$ radians around the origin of the complex plane, and only affects the phases of the entity embeddings in the complex vector space. This idea is motivated by Euler’s identity $e^{i\theta} = cos\theta + isin\theta$, which indicates that a unitary complex
number can be regarded as a rotation in the complex plane.

We regard the relation embedding $\textbf{r}$ as translation from the time-specific subject embedding $\textbf{s}_t$ to the conjugate of the time-specific object embedding $\overline{\textbf{o}}_{t}$ for a single quadruple  (\textit{s}, \textit{r}, \textit{o}, \textit{t})$\in\mathcal{Q^{+}}$, where $\textit{r}\in\mathcal{R}_{b}\cup\mathcal{R}_{e}$ and $\mathcal{Q}^{+}$ denotes the set of all positive quadruples. The score function is defined as:
\begin{equation}
\begin{aligned}\label{scorefunction}
    &f_{\mathrm{TeRo}}(s,r,o,t) = ||\textbf{s}_{t}+\textbf{r} -\overline{\textbf{o}}_{t}||
\end{aligned}
\end{equation}

For a fact (\textit{s}, \textit{r}, \textit{o}, \textit{t}) occurring in a certain time interval, i.e., \textit{t} = [\textit{t}$_b$, \textit{t}$_e$] where $t_{b}, t_{e}$ denote the beginning time and the end time of the fact, we separate this fact into two quadruples, namely, (\textit{s}, \textit{r}$_b$, \textit{o}, \textit{t}$_b$) and (\textit{s}, \textit{r}$_e$, \textit{o}, \textit{t}$_e$). Here, we extend the relation set $\mathcal{R}$ in a TKG which involves time intervals to a pair of dual relation sets, $\mathcal{R}_b$ and $\mathcal{R}_e$. A relation $r_{b}\in\mathcal{R}_{b}$ is used to handle the beginning of relation $r$, meanwhile a relation $r_{e}\in\mathcal{R}_{e}$ is used to handle the end of relation $r$. By doing this, we score a fact (\textit{s}, \textit{r}, \textit{o}, [\textit{t}$_r$, \textit{t}$_e$]) as the mean value of scores of two quadruples, (\textit{s}, \textit{r}$_b$, \textit{o}, \textit{t}$_b$) and (\textit{s}, \textit{r}$_e$, \textit{o}, \textit{t}$_e$) which represent the beginning and the end of this fact respectively.
\begin{align}
     f_{\mathrm{TeRo}}(s,r,o,[t_{b},t_{e}]) = &\frac{1}{2}(||\textbf{s}_{t_{b}}+\textbf{r}_{b} -\overline{\textbf{o}}_{t_{b}}||+||\textbf{s}_{t_{e}}+\textbf{r}_{e}\ -\overline{\textbf{o}}_{t_{e}}||)
\end{align}
Specially, for a fact missing either the beginning time or the end time, e.g., (\textit{s}, \textit{r}, \textit{o}, [\textit{t}$_b$, -]) or (\textit{s}, \textit{r}, \textit{o}, [-, \textit{t}$_e$]), the score of this fact is equal to the score of the quadruple involving the known time, i.e., $f_{\mathrm{TeRo}}(s,r,o,[t_{b},-])=f_{\mathrm{TeRo}}(s,r_{b},o,t_{b})$, $f_{\mathrm{TeRo}}(s,r,o,[-,t_{e}])=f_{\mathrm{TeRo}}(s,r_{e},o,t_{e})$.

In this paper, we use the same loss function as the negative sampling loss proposed in~\cite{RotatE} for optimizing our model. This loss function has been proved to be very effective on optimizing some distance-based KGE models, e.g., TransE, RotatE~\cite{RotatE} and ATiSE~\cite{ATiSE}.
\begin{align}
    L(\xi) =& -\text{log}\ \sigma(\gamma-f_{\mathrm{TeRo}}(\xi))-\sum_{i=1}^{\eta}\frac{1}{\eta}\text{log}\ \sigma(f_{\mathrm{TeRo}}(\xi_{i}^{'})-\gamma)
\end{align}
where $\xi\in\mathcal{Q^{+}}$ is a positive training quadruple, $\xi_{i}^{'}$ is the $i$th negative sample corresponding to $\xi$ generated by randomly corrupting the subject or the objects of $\xi$ such as $(s',p,o,t)$ and $(s,p,o',t)$, $\sigma(\cdot)$ denotes the sigmoid function, $\gamma$ is a fixed margin and $\eta$ is the ratio of negatives over positive training samples. 

\subsection{Learning Various Relation Patterns}

Static KGE models and some existing TKGE models which are the temporal extensions of TransE or DistMult have limitations on capturing some key relation patterns which are defined as follows.
\paragraph{Definition 1.} \textit{A relation r is a \textbf{temporary} relation if $\ \exists s,o,t_{1},t_{2}\ \    r(s,o,t_{1})\land \neg r(s,o,t_{2})$  holds True}
\paragraph{Definition 2.} \textit{A relation r is \textbf{asymmetric} if $\exists s,o,t\ \ r(s,o,t)\land\neg r(o,s,t)$ holds True.}
\paragraph{Definition 3.} \textit{A relation r is a \textbf{reflexive} relation if $\ \exists s,t\ \   r(s,s,t)$ holds True.}

\paragraph{}As mentioned in Section~\ref{Related Work}, static KGE models can not model temporary relations, e.g., '\textit{visits}', since $f_{\mathrm{S KGE}}(s,r,o,t_{1}) \equiv f_{\mathrm{S KGE}}(s,r,o,t_{2})$. Temporal extensions of DistMult (denoted as T-DistMult) including TDistMult, TA-DistMult and Know-Evolve can not model asymmetric relations, e.g., '\textit{parentOf}', since $f_{\mathrm{T-DistMult}}(s,r,o,t)=\langle \textbf{s}_{t},\textbf{r}_{t},\textbf{o}_{t}\rangle= f_{\mathrm{T-DistMult}}(o,r,s,t)$, where \textbf{s}$_{t}$, \textbf{o}$_{t}$, \textbf{r}$_{t}$ are time-specific entity/relation embeddings corresponding to different T-DistMult models.
Temporal extensions of TransE (denoted as T-TransE) including HyTE, TTransE, TA-TransE have difficulties of modeling multiple reflexive relations, e.g., '\textit{equalTo}' and '\textit{subsetOf}', since $f_{\mathrm{T-TransE}}(s_{1},r_{1},s_{1},t)=f_{\mathrm{T-TransE}}(s_{2},r_{2},s_{2},t)=0\Rightarrow \textbf{r}_{1}=\textbf{r}_{2}=0$.

\begin{figure}[h!]
\centering
\vspace{-0.2cm}
\includegraphics[width=\textwidth]{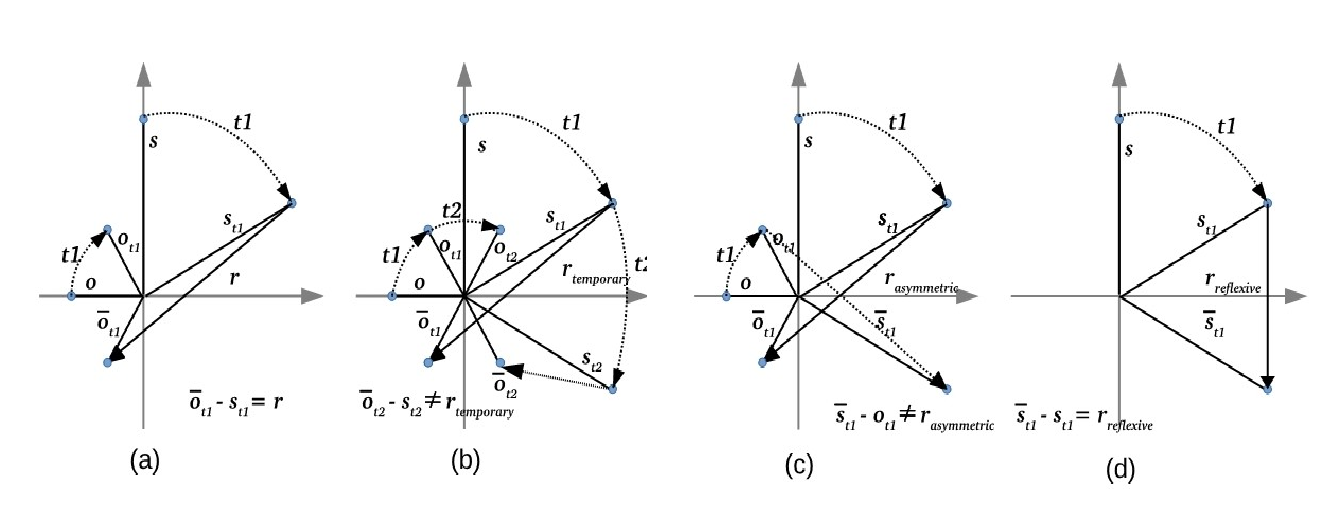} 
\caption{(a) Illustration of TeRo with only one embedding dimension; (b) an example of modeling a temporary relation; (c) an example of modeling an asymmetric relation; (d) an example of modeling a reflexive relation.}
\label{fig:TeRo}
\vspace{-0.2cm}
\end{figure}

By defining each time step as a rotation in the complex vector spaces, TeRo can capture all of the above three relation patterns. Given an observed fact (\textit{s}, \textit{r}, \textit{o}, \textit{t}$_1$) where $\textbf{s}_{t_{1}}+\textbf{r}=\overline{\textbf{o}}_{t_{1}}$:
\begin{itemize}
\setlength{\itemsep}{0pt}

\setlength{\parsep}{0pt}

\setlength{\parskip}{0pt}
    \item as shown in Figure~\ref{fig:TeRo}(b), if $r$ is a temporary relation, we can have $\textbf{s}_{t_{2}}+\textbf{r} \neq \overline{\textbf{o}}_{t_{2}}$ for TeRo to make $r(s,o,t_{1})\land \neg r(s,o,t_{2})$  hold true.
    \item as shown in Figure~\ref{fig:TeRo}(c), if $r$ is an asymmteric relation, we can have $\textbf{o}_{t_{1}}+\textbf{r} \neq \overline{\textbf{s}}_{t_{1}}$ for TeRo to make $r(s,o,t_{1})\land \neg r(o,s,t_{1})$  hold true.
    \item as shown in Figure~\ref{fig:TeRo}(d), if $r$ is a reflexive relation, we have $Im(\textbf{r}) = 2Im(\textbf{s}_{t_{1}})$ for TeRo. Thus, TeRo can represent multiple reflexive relations as different embeddings due to the conjugate operations of object embeddings.
\end{itemize}

\subsection{Complexity}~\label{Complexity}
In Table~\ref{tb:complexity}, we summarize the scoring functions and the space complexites of several state-of-the-art TKGE approaches and our model as well as TransE. $n_e$, $n_r$, $n_{\tau}$ and $n_{token}$ are numbers of entities, relations, time steps and temporal tokens used in~\cite{TA-TransE}; $d$ is the dimensionality of embeddings. $\mathcal{P}_{t}(\cdot)$ denotes the temporal projection for embeddings~\cite{HyTE}. $\textsc{LSTM}(\cdot)$ denotes an LSTM neural network; $[\textbf{r};\textbf{t}_{seq}]$ denotes the concatenation of the relation embedding and the sequence of temporal tokens~\cite{TA-TransE}.  $\overrightarrow{}$ and $\overleftarrow{}$ denote the temporal part and untemporal part of a time-specific diachronic entity embedding~\cite{DE-SimplE}; ${\textbf{r}}^{-1}$ denotes the inverse relation embedding of ${r}$, i.e., $(s,r,o,t)\leftrightarrow (o,{r}^{-1},s,t)$. $\mathcal{D_{KL}}(\cdot)$ denotes the KL divergence between two Gaussian distributions; $\mathbf{P}_{s,t}, \mathbf{P}_{o,t},\mathbf{P}_{r,t}$ denote the Gaussian embeddings of $s$, $r$ and $o$ at time $t$~\cite{ATiSE}.

As shown in Table~\ref{tb:complexity}, the space complexity of TeRo and TransE will be close if $n_{\tau}<n_{e}$. In practice, we can achieve this condition by tuning the time granularity.

\begin{table}[h!]
\centering
\begin{tabular}{|l|c|c|}
\hline
Model&Scoring Function&Space Complexity\cr
\hline
TransE&$||\textbf{s}+\textbf{r}-\textbf{o}||$&$\mathcal{O}(n_{e}d+n_{r}d)$\cr
TTransE&$||\textbf{s}+\textbf{r}+\boldsymbol{\tau}-\textbf{o}||$&$\mathcal{O}(n_{e}d+n_{r}d+n_{\tau}d)$\cr
HyTE&$||P_{t}(\textbf{s})+P_{t}(\textbf{r})-P_{t}(\textbf{o})||$&$\mathcal{O}(n_{e}d+n_{r}d+n_{\tau}d)$\cr
TA-TransE&$||\textbf{s}+\textsc{LSTM}([\textbf{r};\textbf{t}_{seq}])-\textbf{o}||$&$\mathcal{O}(n_{e}d+n_{r}d+n_{token}d)$\cr
TA-DistMult&$\langle \textbf{s},\textsc{LSTM}([\textbf{r};\textbf{t}_{seq}]),\textbf{o}\rangle$&$\mathcal{O}(n_{e}d+n_{r}d+n_{token}d)$\cr
DE-SimplE&$\frac{1}{2}(\langle\overrightarrow{\textbf{s}}^{t},\textbf{r},\overleftarrow{\textbf{o}}^{t}\rangle+\langle\overrightarrow{\textbf{o}}^{t},\textbf{r}^{-1},\overleftarrow{\textbf{s}}^{t}\rangle)$&$\mathcal{O}(n_{e}d+n_{r}d)$\cr
ATiSE&$\mathcal{D_{KL}}(\mathbf{P}_{s,t}-\mathbf{P}_{o,t},\mathbf{P}_{r,t})$&$\mathcal{O}(n_{e}d+n_{r}d)$\cr
\hline
TeRo&$||\textbf{s}_{t}+\textbf{r} -\overline{\textbf{o}}_{t}||$&$\mathcal{O}(n_{e}d+n_{r}d+n_{\tau}d)$\cr
\hline
\end{tabular}
\caption{Comparison of our models with several baseline models for space complexity.}
\label{tb:complexity}
\vspace{-0.3cm}
\end{table}

\section{Experiments}

\subsection{Temporal Knowledge Graph Datasets}
Common TKG benchmarks include GDELT~\cite{DE-SimplE}, ICEWS14, ICEWS05-15, YAGO15k, Wikidata11k~\cite{TA-TransE}, YAGO11k and Wikidata12k~\cite{HyTE}. In this work, we choose ICEWS14, ICEWS05-15, YAGO11k and Wikidta12k as datasets for the following reasons: 1. ICEWS14 and ICEWS05-15 are two well-established event-based datasets which are commonly used in previous literature~\cite{TA-TransE,DE-SimplE,ATiSE}, these two datasets are subsets of ICEWS~\cite{ICEWS2015} corresponding to facts in 2014 and facts between 2005 and 2015, where all time annotations are time points; 2. YAGO15k, Wikidata11k, YAGO11k and Wikidata12k are subsets of YAGO3~\cite{YAGO3} and Wikidata~\cite{Wikidata} where a part of time annotations are time intervals. In YAGO15k and Wikidata11k, each time interval only contains either beginning dates or end dates, shaped like ’\textit{occurSince 2003}’ or ’\textit{occurUntill 2005}’ and a part of facts in YAGO15k exclude time information. Thus we prefer to using YAGO11k and Wikidata12k where each fact includes time information and time annotations are represented in the various forms, i.e., time points like [\textit{2003-01-01}, \textit{2003-01-01}], beginning or end time like [\textit{2003}, \textit{\#\#}], and time intervals like [\textit{2003}, \textit{2005}]. We list the statistics of the four datasets we use in Table~\ref{tb:dataset}.
\begin{table}[h!]
\centering
\resizebox{0.8\textwidth}{!}{
\begin{tabular}{|l|c|c|c|c|c|c|}
  \hline
  Dataset & \#Entities &\#Relations &Time Span& \#Training&\#Validation&\#Test\\
  \hline
  ICEWS14 & 6,869 & 230 & 2014&72,826&8,941&8,963 \\
  ICEWS05-15&10,094&251&2005-2015&368,962&46,275&46,092\\
  YAGO11k&10,623 &10 & -453-2844&16,406  &2,050 &2,051\\
  Wikidata12k&12,554 &24 &1479-2018&32,497&4,062&4,062\\
  \hline
\end{tabular}}
\caption{Statistics of datasets.
  }\label{tb:dataset}
\vspace{-0.1cm}
\end{table}

\subsection{Time Granularity}~\label{time granularity}
 In some recent work~\cite{HyTE,ATiSE}, the time span of a TKG dataset was splitted into a number of time steps. For ICEWS14 and ICEWS05-15, the time granularity was fixed as 1 day. For YAGO11k and Wikidata12k, month and day information was dropped, and less frequent year mentioned were clubbed into same time steps but years with high frequency formed individual time steps in order to alleviate the effect of the long-tail property of time data. In other words, the lengths of different time steps were different for the balance of numbers of triples in different time steps. However, it has not been investigated whether the lengths of time steps affect the performances of TKGE models. 


In this work, we test our model with different time units, denoted as $u$, in a range of \{1, 2, 3, 7, 14, 30, 90 and 365\} days for ICEWS datasets. Dasgupta et al.~\shortcite{HyTE} and Xu et al.~\shortcite{ATiSE} applied a minimum threshold
of 300 triples per interval during construction for YAGO11k and Wikidata12k. We follow their time-division approaches for these two datasets and test different minimum thresholds, denoted as $thre$, in a range of \{1, 3, 10, 30, 100, 300, 1000, 3000, 10000, 30000\}. The change of time granularity will reconstruct the set of time steps $\mathcal{T}$. For ICEWS14, when the time unit $u$ is 1 day, we have totally 365 time steps and the date \textit{2014-01-02} is represented by the second time step, i.e., $\tau_{1}$. If the time unit is changed as 2 days, the total number of time steps will be 183 and the date \textit{2014-01-02} will be denoted as $\tau_{0}$. For YAGO11k, when the mini threshold $thre=1$, we have 396 time steps since there are totally 396 
different years existing as timestamps in YAGO11k. Years like \textit{-453}, \textit{100} and \textit{2008} are all taken as independent time steps. When $thre$ for YAGO11k rises to 300, the number of time steps drops to 127 and years between \textit{-431} and \textit{100} are clubbed into a same time step.
\subsection{Evaluation Metrics}
We evaluate our model by testing the performances of our model on link prediction task over TKGs under the time-wise filtered setting defined in~\cite{ATiSE,DE-SimplE}. This task is to complete a time-wise fact with a missing entity. For a test quadruple $(s, p, o, t)$, we first generate candidate quadruples $\mathcal{C}=\{(s, p, o', t):o'\in \mathcal{E}\}\cup\{(s', p, o, t):s'\in \mathcal{E}\}$ by replacing $s$ or $o$ with all possible entities. Different from the time-unwise filtered setting~\cite{TransE} which filters the triples appearing either in the training, validation or test set from the candidate list , we only filter the quadruples $\xi\in\mathcal{Q^{+}}$ existing in the dataset. This ensures that the facts which do not appear at time $t$ are still considered as candidates for evaluating the given test quadruple. We obtain the final rank of the test quadruple among filtered candidate quadruples $\overline{\mathcal{C}}=\{\xi:\xi\in\mathcal{C},\xi\notin\mathcal{Q^{+}}\}$ by sorting their scores.

Two commonly used evaluation metrics are used here, i.e., Mean Reciprocal Rank and Hits@k. The Mean Reciprocal Rank (MRR) is the means of the reciprocal values of all computed ranks. And the fraction of test quadruples ranking in the top $k$ is called Hits@k. 

\subsection{Baselines}
We compare our approach with several state-of-the-art KGE approaches and existing TKGE approaches, including TransE~\cite{TransE}, DistMult~\cite{DISTMULT}, ComplEx-N3~\cite{Complex-N3}, RotatE~\cite{RotatE}, QuatE~\cite{QuatE}, TTransE~\cite{leblay}, TA-TransE, TA-DistMult~\cite{TA-TransE}, DE-SimplE~\cite{DE-SimplE} and ATiSE~\cite{ATiSE}. The results of most baselines are taken from some recent work~\cite{DE-SimplE,ATiSE} which used the same evaluation protocol as ours. DE-SimplE which mainly focuses on event-based datasets, cannot model time intervals or time annotations missing moth and day information which are common in YAGO and Wikidata. Thus its result on YAGO11k and Wikidata12k are unobtainable.  Since the original source code of TA-TransE and TA-DistMult~\cite{TA-TransE} is not released, we reimplement these models according to the implementation details reported in the original paper, in order to obtain their results on YAGO11k and Wikidata12k.

\subsection{Experimental Setup}
We implement our proposed model in PyTorch. The code is available at \url{https://github.com/soledad921/ATISE}. 

We select the optimal hyperparameters by early validation stopping according to MRR on the validation set. 
We restrict the iterations to 5000. Following the setup used in~\cite{ATiSE},
the batch size $b= 512$ is kept for all datasets, the embedding dimensionality $k$ is tuned in \{$100,200,300,400,500$\}, the ratio of negative over positive training samples $\eta$ is tuned in \{$1,3,5,10$\} and the margin $\gamma$ is tuned in \{1, 2, 3, 5, 10, 20, $\cdots$, 120\}. Regarding optimizer, we choose Adagrad for TeRo and tune the learning rate $r$ in a range of \{$1,0.3,0.1,0.03,0.01$\}. Specially, the time granularity parameters $u$ and $thre$ are also regraded as hyperparameters for TeRo as mentioned in Section~\ref{time granularity}.

The default configuration for TeRo is as follows: $d=500$, $\eta=10$. Below, we only list the non-default parameters: $lr=0.1$, $\gamma=110$, $u=1$ on ICEWS14; $lr=0.1$,  $\gamma=120$, $u=2$ on ICEWS05-15; $lr =0.1$, $\gamma=50$, $thre=100$ on YAGO11k; $lr =0.3$, $\gamma=20$, $thre=300$ on Wikidata12k.

\section{Results and Analysis}
\subsection{Comparative Study}
\begin{table*}[h!]
\centering
\resizebox{0.9\textwidth}{!}{
\begin{tabular}{|c|c|c|c|c|c|c|c|c|}
    \hline
  Datasets&\multicolumn{4}{c|}{ICEWS14}&\multicolumn{4}{c|}{ICEWS05-15}\cr 
       \hline   
           Metrics&MRR&Hits@1&Hits@3&Hits@10&MRR&Hits@1&Hits@3&Hits@10\cr
\hline
 TransE*  &.280 &.094 &- &.637 &.294 &.090 &- &.663 \cr
        DistMult* &.439 &.323 &- &.672 &.456 &.337 &- &.691 \cr
        ComplEx-N3$^{\dagger}$  &.467 &.347 &.527 &.716 &.481 &.362 &.535 &.729  \cr
        RotatE$^{\dagger}$ &.418 &.291 &.478 &.690 &.304 &.164 &.355 &.595 \cr
        QuatE$^{2}$ $^{\dagger}$ &.471 &.353 &.530 &.712 &.482 &.370 &.529 &.727 \cr
\hline
        TTransE$^{\diamond}$&.255 &.074 &- &.601 &.271 &.084 &- &.616 \cr
        HyTE$^{\diamond}$ &.297 &.108 &.416  &.655 &.316  &.116  &.445  &.681 \cr
        TA-TransE* &.275  &.095
        &-
        &.625 &.299 &.096 &- &.668  \cr
        TA-DistMult* &.477 &.363  &-  &.686  &.474 &.346  &- &.728   \cr
        DE-SimplE$^{\diamond}$ &.526  &.418 &.592  &.725 &.513  &.392  &.578  &.748 \cr
        ATiSE$^{\dagger}$ &.550 &.436 &\textbf{.629} &\textbf{.750} &.519 &.378 &.606 &.794\cr
        \hline
        TeRo&\textbf{.562} &\textbf{.468} &.621 &.732 &\textbf{.586} &\textbf{.469} &\textbf{.668} &\textbf{.795}\cr
\hline
 
\end{tabular}}
\caption{
    Link prediction results on ICEWS14 and ICEWS05-15. 
    *: results are taken from~\cite{TA-TransE}. $^{\diamond}$: results are taken from~\cite{DE-SimplE}. $^{\dagger}$: results are taken from~\cite{ATiSE}. Dashes: results are unobtainable.
     The best results among all models are written bold.
    }
    \label{ICEWS results}
\end{table*}

\begin{table*}[h!]
\centering
\resizebox{0.9\textwidth}{!}{
\begin{tabular}{|c|c|c|c|c|c|c|c|c|}
    \hline
  Datasets&\multicolumn{4}{c|}{YAGO11k}&\multicolumn{4}{c|}{Wikidata12k}\cr 
       \hline   
           Metrics&MRR&Hits@1&Hits@3&Hits@10&MRR&Hits@1&Hits@3&Hits@10\cr
\hline
 TransE$^{\dagger}$  &.100 &.015 &.138 &.244 &.178 &.100 &.192 &.339 \cr
        DistMult$^{\dagger}$ &.158 &.107 &.161 &.268 &.222 &.119 &.238 &.460 \cr
        ComplEx-N3$^{\dagger}$  &.167 &.106 &.154 &.282 &.233 &.123 &.253 &.436  \cr
        RotatE$^{\dagger}$ &.167 &.103 &.167 &.305 &.221 &.116 &.236 &.461 \cr
        QuatE$^{2}$ $^{\dagger}$ &.164 &.107 &.148 &.270 &.230 &.125 &.243 &.416 \cr
\hline
        TTransE$^{\dagger}$&.108 &.020 &.150 &.251 &.172 &.096 &.184 &.329 \cr
        HyTE$^{\dagger}$ &.105  &.015 &.143  &.272 &.180  &.098 &.197  &.333 \cr
        TA-TransE &.127  &.027 &.160 &.326 &.178  &.030 &.267  &.429 \cr
        TA-DistMult &.161  &.103 &.171  &.292 &.218  &.122 &.232  &.447 \cr
        ATiSE$^{\dagger}$ &.170 &.110 &.171 &.288 &.280 &.175 &.317 &.481\cr
        \hline
        TeRo&\textbf{.187} &\textbf{.121} &\textbf{.197} &\textbf{.319} &\textbf{.299} &\textbf{.198} &\textbf{.329} &\textbf{.507} \cr
\hline
\end{tabular}}
\caption{
    Link prediction results on YAGO11k and Wikidata12k. $^{\dagger}$: results are taken from~\cite{ATiSE}. The best results among all models are written bold.
    }
    \label{YAGO results}
\end{table*}

Table~\ref{ICEWS results} and~\ref{YAGO results} list all link prediction results of our proposed model and baseline models on four datasets. TeRo surpassed all baseline embedding models regarding all metrics on all datasets except that the ATiSE got the better Hits@3 and Hits@10 than TeRo on ICEWS14. Compared to ATiSE, TeRo achieved the improvement of 1.2 MRR points, 6.7 MRR points, 1.7 MRR points and 1.9 MRR points on ICEWS14, ICEWS05-15, YAGO11k and Wikidata12k respectively.
\subsection{Ablation Study}
\begin{figure}[h!]
    \centering
        \vspace{-0.1cm}
    \includegraphics[width=0.9\textwidth]{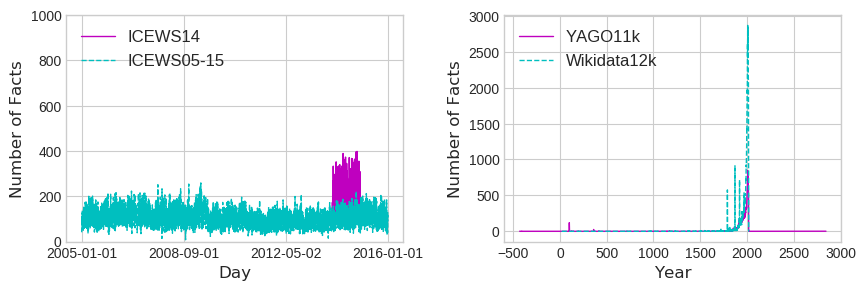} 
    \vspace{-0.6cm}
    \caption{Time distribution of numbers of facts.}
    \label{fig:timedistribution}
    \vspace{-0.3cm}
    \end{figure}

In this work, we analyze the effect of the change of the time granularity on the performance of our model. As mentioned in Section~\ref{time granularity}, we adopt two different time-division approaches for event-based datasets, i.e., ICEWS datasets, and time-wise KGs involving time intervals, i.e., YAGO11k as well as Wikidata12k. For ICEWS14 and ICEWS05-15, we use time steps with fixed length since the the distribution of numbers of facts in ICEWS datasets over time are relatively uniform as shown in Figure~\ref{fig:timedistribution}. The time granularities of ICEWS datasets are equal to the lengths of time units $u$ . On the other hand, the time distributions of numbers of facts in YAGO15k and Wikidata12k are long-tailed. Thus we divide the time steps in YAGO15k and Wikidata12k by setting a mini threshold for the numbers of facts in each time step. Time granularities of these two datasets can be changed by setting different thresholds $thre$.
    
    \begin{figure}[h!]
    \centering
    \includegraphics[width=0.9\textwidth]{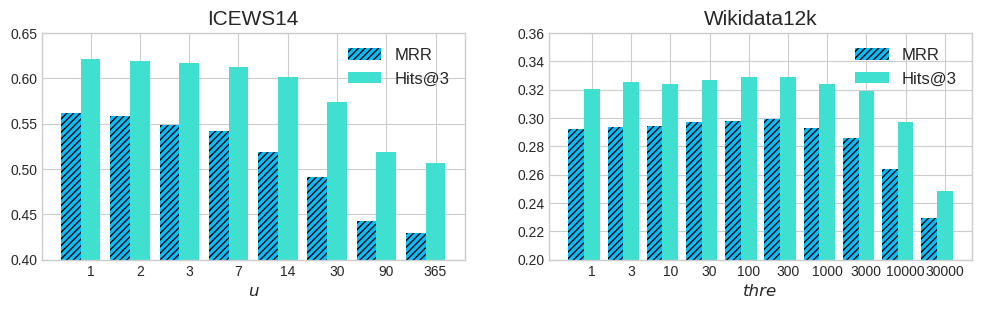} 
    \vspace{-0.6cm}
    \caption{Results of TeRo with different time granularities on ICEWS14 and Wikidata12k.}
    \label{fig:time granularity}
    \vspace{-0.15cm}
    \end{figure}

In ICEWS14, time distribution is relatively uniform and thus representing time with a small time granularity can provide more abundant time information. As shown in Figure~\ref{fig:time granularity}, TeRo with small time granularities, e.g., 1 day, 2 days and 3 days, had better performance on ICEWS14 compared to TeRo with big time granularities regarding MRR and Hits@3. Likewise, the optimal time unit for TeRo on ICEWS05-15 was proven by our experiments to be 2 days. For Wikidata12k, using a very small time granularity was non-optimal due to the long-tail property of time data. On the other hand, using an overly big time granularity resulted in the invalid incorporation of time information. Figure~\ref{fig:time granularity} demonstrates the low performances of TeRo with big time granularities. More concretely, when time unit $u$ was 1 year, all of time annotations in ICEWS14 were represented by a uniform time embedding, which meant this time embedding was temporally unmeaningful. Table~\ref{tb:link prediction} demonstrates a few examples of link prediction results on ICEWS14 of TeRo models with time units $u$ of two days and one year.
\begin{table*}[h]
\centering
\resizebox{\textwidth}{!}{ 
\begin{tabular}{|l|l|l|}
  \hline
 Link Prediction&TeRo with $u=1$ day&TeRo with $u=365$ days\cr
  \hline
Colombia, Host a vist, ?, 2014-06-04&\textbf{Kyung-wha Kang} &John F. Kelly \cr
Head of Government (China), visits, ?, 2014-07-04 &\textbf{South Korea}&Serbia \cr
UN Security Council, Criticize or denounce, ?, 2014-08-10 &\textbf{North Korea}&Armed Band (South Sudan)\cr
South Korea, Host a vist, ?, 2014-06-20 &\textbf{Kim Jong-Un}&National Security Advisor (Japan)\cr
Police (Australia), Accuse, ?, 2014-10-22&\textbf{Criminal (Australia)} &Citizen (Australia)\cr
    
  \hline
\end{tabular}
}    \caption{
Examples of link prediction results on ICEWS14. The correct predictions are written \textbf{bold}.
    }    \label{tb:link prediction}
    \vspace{-0.05cm}
\end{table*}

As shown in Table~\ref{tb:link prediction}, in many cases, TeRo with $u=1$ predicated correctly, meanwhile TeRo with $u=365$ gave the wrong predictions. We notice that these predictions of TeRo with $u=365$ in Table~\ref{tb:link prediction} would be valid if we disregarded the time constraint. For instance, (\textit{Colombia}, \textit{Host a visit}, \textit{John F. Kelly}) happened on 2014-03-27, (\textit{UN Security Council}, \textit{Criticize or denounce}, \textit{Armed Band (South Sudan)}) was true on 2014-08-07. As mentioned in Section~\ref{method}, \textit{Host a visit} and \textit{Criticize or denounce} are temporary relations. The above results prove that using a reasonable time granularity is helpful for TeRo to effectively incorporate time information. And the inclusion of time information enables TeRo to capture temporary relations and improve its performance on link prediction over TKGs.
\subsection{Efficiency Study}
TeRo has the same space complextiy as TTransE~\cite{leblay} and HyTE~\cite{HyTE}. Since we constrained the numbers of time steps of the four TKG datasets by tuning time granularities (183 time steps in ICEWS14, 1339 time steps in ICEWS05-15, 127 time steps in YAGO11k and 82 time steps in Wikidata12k), the numbers of time steps are much less than the numbers of entities in these datasets, which means that the space complexity of TeRo is close to the space complexity of TransE~\cite{TransE} as mentioned in Section~\ref{Complexity}. Regarding the concrete memory consumption, the recent state-of-the-art TKGE models, ATiSE~\cite{ATiSE} and DE-SimplE~\cite{DE-SimplE} have 1.8 times and 2.2 times as large 
memory size as TeRo on ICEWS14 with the same embedding dimensionality. The training processes of TeRo with 500-dimensional embeddings on ICEWS14, ICEWS05-15, YAGO11k and Wikidata12k take 4.3 seconds, 25.9 seconds, 1.9 seconds and 4.1 seconds per epoch on a single GeForce RTX 2080 device, respectivly.
\begin{figure}[h!]
\centering
\includegraphics[width=0.7\textwidth]{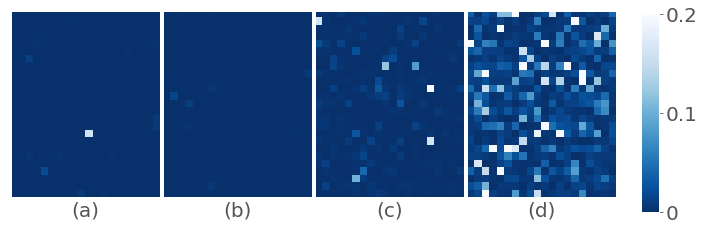} 
\vspace{-0.2cm}
\caption{Visualization of the absolute difference vectors between $\textbf{r}_b$ and $\textbf{r}_e$ for relations \textit{deadIn} and \textit{isMarriedTo} (reshaped into 25$\times$20 matrices): (a) $|Re(\textbf{r}_{b}-\textbf{r}_{e})|/|Re(\textbf{r}_{b})|$ for relation \textit{deadIn}; (b) $|Im(\textbf{r}_{b}-\textbf{r}_{e})|/|Im(\textbf{r}_{b})|$ for relation \textit{deadIn}; (c) $|Re(\textbf{r}_{b}-\textbf{r}_{e})|/|Re(\textbf{r}_{b})|$ for relation \textit{isMarriedTo}; (d) $|Im(\textbf{r}_{b}-\textbf{r}_{e})|/|Im(\textbf{r}_{b})|$ for relation \textit{isMarriedTo}.}
\label{fig:RelationEmbedding}
\vspace{-0.2cm}
\end{figure}

It is also noteworthy that representing each relation as a pair of dual complex embeddings is helpful to save training time on TKGs involving time intervals. Given a fact ($s$, $r$, $o$, [$t_b$, $t_e$]), some TKGE models, e.g., HyTE and ATiSE, discretize this fact into several quadruples involving continuous time points, i.e., [($s$, $r$, $o$, $t_b$), ($s$, $r$, $o$, $t_{b}+1$), $\cdots$, ($s$, $r$, $o$, $t_e$)]. When $thre=300$, each fact lasts for averagely around 15 and 8 time steps in YAGO11k and Wikidata12k. In other words, such method that discretizes facts involving time intervals expands the sizes of both datasets by 15 and 8 times. In our model, we propose a more efficient method to handle time intervals by using two different quadruples, ($s$, $r_b$, $o$, $t_b$) and ($s$, $r_e$, $o$, $t_b$) to represent the beginning and the end of each fact. In this way, we only expand the sizes of datasets as less than twice as their original sizes. 

For relations $r$ in YAGO11k, we analyze the similarities between the embeddings $\mathbf{r}_b$ and $\mathbf{r}_e$. As shown in Figure~\ref{fig:RelationEmbedding}, for short-term relations, e.g., \textit{deadIn}, the real parts of $\textbf{r}_b$ and $\textbf{r}_e$, as well as their imaginary parts, have high similarities since ${r}_b$ and ${r}_e$ always happen at the same time and have the same semantics. By contrast, for long-term relations, e.g., \textit{isMarriedTo}, the real parts of $\textbf{r}_b$ and $\textbf{r}_e$ show their semantic similarities and the imaginary parts capture their temporal dissimilarities.

\section{Conclusion}
In this work, we introduce TeRo, a new TKGE model which represents entities or relations as single or dual complex embeddings and temporal changes as rotations of entity embeddings in the complex vector space. Our model is advantageous with its capability in modelling several key relation patterns and handling time annotations in various forms. Experimental results show that TeRo remarkably outperforms the existing state-of-the-art KGE models and TKGE models on link prediction over four well-established TKG datasets. Specially, we adopt two different time-division approaches for various datasets and investigate the effect of the time granularity on the performance of our model.
\section*{Acknowledgements}
This work is supported by the CLEOPATRA project (GA no.~812997), the German national funded BmBF project MLwin and the BOOST project.
\bibliographystyle{coling}
\bibliography{coling2020}

\end{document}